\documentclass{article}

\usepackage{microtype}
\usepackage{graphicx}
\usepackage{subcaption}
\usepackage{booktabs} 

\usepackage{hyperref}

\usepackage[preprint]{icml2026/icml2026}

\usepackage{amsmath}
\usepackage{amssymb}
\usepackage{mathtools}
\usepackage{amsthm}
\usepackage{bm}

\usepackage[capitalize,noabbrev]{cleveref}

\theoremstyle{plain}

\theoremstyle{definition}

\theoremstyle{remark}

\usepackage[textsize=tiny]{todonotes}

\icmltitlerunning{Jigsaw Regularization in Whole-Slide Image Classification}

\begin{document}

\twocolumn[
  \icmltitle{Jigsaw Regularization in Whole-Slide Image Classification}



  \icmlsetsymbol{equal}{*}

  \begin{icmlauthorlist}
    \icmlauthor{So Won Jeong}{yyy}
\icmlauthor{Veronika Ročková}{yyy}
    
  \end{icmlauthorlist}

  \icmlaffiliation{yyy}{Booth School of Business, The University of Chicago, IL, USA}


  \icmlcorrespondingauthor{So Won Jeong}{sowonjeong@chicagobooth.edu}
  \icmlkeywords{Machine Learning, ICML}
  \vskip 0.3in
]



\printAffiliationsAndNotice{}  

\begin{abstract}
Computational pathology involves the digitization of stained tissues into whole-slide images (WSIs) that contain billions of pixels arranged as contiguous patches. 
Statistical analysis of WSIs largely focuses on classification via multiple instance learning (MIL), in which slide-level labels are inferred from unlabeled patches. 
Most MIL methods treat patches as exchangeable, overlooking the rich spatial and topological structure that underlies tissue images.
This work builds on recent graph-based methods that aim to incorporate spatial awareness into MIL.
Our approach is new in two regards: (1) we deploy vision \emph{foundation-model embeddings}
to incorporate local spatial structure within each patch, and (2) achieve across-patch spatial awareness using graph neural networks together with a novel {\em jigsaw regularization}. 
We find that a combination of
these two features markedly improves classification over state-of-the-art attention-based MIL approaches on benchmark datasets in breast, head-and-neck, and colon cancer.
\end{abstract}

\section{Introduction}

\begin{figure*}[!t]
    \centering
    \includegraphics[width=0.9\linewidth]{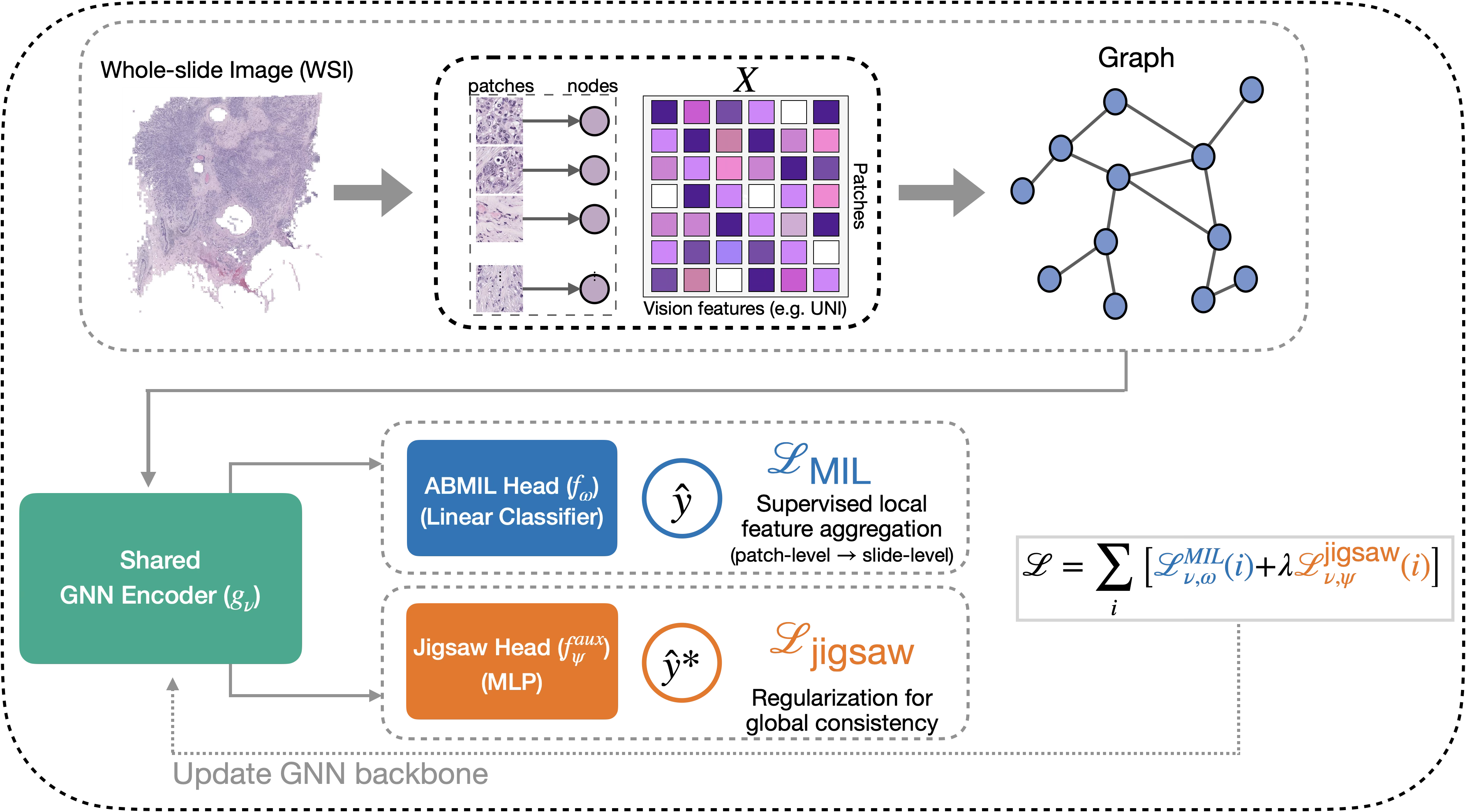}
    \caption{Jigsaw-regularized graph ABMIL framework. There is a shared encoder backbone (green).}
    \label{fig:jigsaw-gnn-framework}
\end{figure*}

Whole-slide images (WSIs) are high-resolution digitizations of histopathology slides and constitute a key data asset in computational pathology.
Owing to their massive scale (billions of pixels), WSIs generally cannot be processed holistically.
Instead, each image slide is typically parsed patch-by-patch by tiling the slide into thousands of smaller regions for downstream analysis \citep{chen2024towards}.
While patch-based processing enables scalable learning, it weakens the connection between representation and supervision. 
In most clinical settings, the outcome of interest is defined at the slide or patient level, whereas learning is performed on collections of unlabeled patches. 
As a result, models must aggregate information from many local observations to infer a global decision, often under weak supervision.

To decouple representation learning from downstream inference, many state-of-the-art computational pathology models deploy a two-step strategy \citep{lu2021data}.
First, a pretrained vision foundation model is used to extract patch-level embeddings \citep{lu2021data, chen2024towards}. 
These foundation models operate in a manner that is very similar to language models, learning representations via self-supervised training objectives such as masked modeling
or contrastive learning \citep{dosovitskiy2021an, oquab2023dinov2}.
Second, a task-specific aggregation model combines these embeddings to produce slide-level predictions. 
This step is most often handled by Multiple Instance Learning (MIL), a weakly supervised learning framework in which labels are available only for bags of observations (entire slides) rather than for individual instances (patches). 
The main engine behind contemporary MIL implementations is hierarchical deep learning classification that is often endowed with additional features such as attention-weighted pooling \citep{ilse2018attention}. 
Attention-based Multiple Instance Learning (ABMIL) \citep{ilse2018attention} introduces a learnable attention mechanism that attaches an ``attention score'' to each instance which enables localization of patches with most classification influence. 
Aggregations of MIL classifiers though ensembling
has received considerable  success as well \citep{campanella2019clinical}.
Recent foundation-model-driven pipelines that pair pre-trained pathology encoders with weakly supervised aggregation also include \citet{lu2021data,ahn2024histopathologic, wang2024pathology}.

Early MIL approaches treat patches as \emph{exchangeable}, effectively ignoring any relationships or ordering among them \citep{ilse2018attention}.
This treatment is poorly aligned with histopathological practice, where diagnostic cues are often detected from certain spatial tissue arrangements. 
Recent work has sought to relax this assumption by incorporating local spatial context, for example through graph-based MIL formulations \citep{li2018graph, lu2020capturing, mesquita2020rethinking, zhao2020predicting, chen2021whole,ma2024gcn} or segmentation-guided aggregation \citep{fang2024sam}. 
Graph-based MIL incorporate Graph Neural Networks (GNNs) that update node embeddings via message passing along a pre-determined graph.
These methods primarily capture short-range dependencies through a fixed network structure based on patch proximity. 
However, global  spatial structure might be missed if the graph fails to connect patches that, while farther away, contain similar tissue.
We defer a more detailed literature review to Appendix~\ref{sec:related_works}.

Our work fortifies spatial awareness within MIL by incorporating both local and global spatial features in (binary) image classification. 
Local spatial relationships between patches are modeled using a graph representation with GNNs. 
Global spatial awareness is encouraged by a jigsaw regularizer that is trained to infer the absolute spatial location of a patch based solely on its GNN-refined embedding.
Jigsaw regularization forces the classifier to answer:  ``Where does this patch belong?"

Our classification head builds on state-of-the-art MIL approaches and consists of several components explained below. See Figure~\ref{fig:jigsaw-gnn-framework} for the overall pipeline.

\vspace{-0.3cm}
\begin{itemize}
    \item \textbf{Foundation-model Embeddings:}
    A pre-trained pathology foundation models \citep{wang2022transformer, chen2024towards, xu2024whole, vorontsov2024foundation} provide fixed patch embeddings which provide high-quality image features that, while invisible to the human eye, have improved prediction capability compared to linear (e.g. wavelet based) pixel aggregates.
   \item \textbf{Graph-MIL:}  
We employ a Graph Multiple Instance Learning framework that models \emph{local} spatial relationships between tissue patches using distance-based $k$ nearest neighbor (NN) graphs and Graph Attention Networks, relaxing the exchangeability assumption inherent in standard MIL.
\item \textbf{Jigsaw Regularization:}  
We enhance Graph-MIL with a regularizer based on a jigsaw puzzle \citep{chen2023jigsaw} where we treat each patch as a piece of puzzle.
This regularizer is based on a self-supervised objective that encourages the model to recover the spatial grid location of each patch, thereby enforcing global structural awareness.
\end{itemize}

\vspace{-0.5cm}
To balance the influence of jigsaw regularization in the training objective relative to the supervised MIL loss, we treat the regularization strength (denoted as $\lambda$) as a latent variable and apply an updating scheme inspired by Expectation-Maximization (EM) algorithm for its tuning.
This ensures to balance two different tasks for training in a reliable manner rather than heuristics \citep{crawshaw2020multi, kendall2018multi}.

We demonstrate consistent improvements over existing MIL baselines across multiple cancer types.
We employ our approach for a variety of pathology foundation models and consistently show improved classification performance.
We conclude that performance gains arise from careful spatial modeling rather than reliance on a specific embedding backbone.
We use several benchmark datasets from The Cancer Genome Atlas Project (TCGA) \citep{weinstein2013cancer} including both balanced and imbalanced samples that exhibit different morphological behavior.
Our experiments show that adding the jigsaw regularizer on top of graph-based MIL (the novel aspect in our work) offers non-negligible improvements across tissue types, foundation models, even for smaller training sample sizes. 
We believe that this enhancement moves digital pathology closer to becoming an integral part of a daily decision routine for clinical pathologists.

\section{Methodology}
In computational pathology, multiple instance learning (MIL) treats each whole-slide image (WSI) as a bag of image patches and learns slide-level predictions by aggregating patch-level representations under weak supervision, without requiring patch-level annotations.

We build on this paradigm by introducing a spatially-aware MIL framework that departs from the standard exchangeability assumption. 
Specifically, we model each WSI as a structured spatial graph, enabling \emph{local} context integration via graph-based message passing, and impose \emph{global} structural coherence through a self-supervised jigsaw regularization task. 
Together, these components allow the model to exploit both local and global within-slide spatial organization while aiming for slide-level inference.

\subsection{MIL in Computational Pathology}\label{sec:mil_in_pathology}
We regard the $i^{th}$ whole-slide image as a bag $B_i = \{x_{i1}, \ldots, x_{i N_i}\}$ of patch-level representations $x_{ij} \in \mathbb{R}^{d_1}$.
While   prior MIL literature was largely focused on the optimization of   pixel-to-feature mappings, often requiring computationally expensive end-to-end training, we bypass this bottleneck by harnessing the representational power of emerging foundational models.
Consequently, $x_{ij}$ denotes \emph{pre-trained} embeddings extracted from the slide using a foundational vision transformer (e.g., the UNI model \citep{chen2024towards}, which generates 1024-dimensional embeddings, i.e., $d_1=1024$).

The associated label $y_i\in\{0,1\}$ corresponds to a slide-level outcome (e.g., tumor presence, molecular status, or survival). 
Instance-level labels $y_{ij}$, indicating whether $j^{th}$ patch is diagnostically relevant, are unobserved.

In the original multiple instance learning  formulation as in \citet{dietterich1997solving}, a bag is labeled positive if at least one of its instances is positive, i.e.
$y_i =1-\prod_{j=1}^{N_i} (1- y_{ij})$. 
Contemporary MIL models in digital pathology have largely moved away from this instance-level probability pooling (which often relies on max pooling) toward embedding-level aggregation \citep{ilse2018attention}.
In this paradigm, the model learns a slide-level prediction $\hat{y}_i$ by first mapping instances to a feature space, and then aggregating these representations through a classifier.

The baseline model expresses the slide-level probability $\mathbb{P}(y_i = 1)$ using a hierarchical model that (1) refines pre-trained patch embeddings $x_{ij}$ using an instance encoder $g_\nu: \mathbb{R}^{d_1} \to \mathbb{R}^{d_2}$, (2) aggregates these patch-level refinements $g_\nu(x_{ij})$ into slide-level embeddings
$$\mathbf{z}_i = \rho_\eta \Big( \{ g_\nu(x_{ij})\}_{j=1}^{N_i} \Big)$$
through an aggregator $\rho_\eta$.
Formally, $\rho_\eta$ maps any finite subset of $\mathbb{R}^{d_2}$ (i.e., a bag of variable size $N_i$) to a fixed-dimensional vector in $\mathbb{R}^{d_3}$.
Finally, (3) it links the slide embedding $\mathbf{z}_i$ to the outcome via logistic regression.

In standard MIL approaches, the refinement $g_\nu(\cdot)$ acts on each patch $x_{ij}$ in isolation. 
It is typically modeled with a Multi-Layer Perceptron (MLP),
\begin{equation*}
    g_\nu(x_{ij}) = \text{ReLU}(\mathbf{W}_2 \cdot \text{ReLU}(\mathbf{W}_1 x_{ij} + \mathbf{b}_1) + \mathbf{b}_2),
\end{equation*}
where the learnable parameters are $\nu = \{\mathbf{W}_1, \mathbf{b}_1, \mathbf{W}_2, \mathbf{b}_2\}$.
Because this function does not contain information from other patches in the bag, the refined embeddings $\{h_{ij}\}_{j=1}^{N_i}$ where $h_{ij} = g_\nu(x_{ij})$ are agnostic to the spatial structure. This work remedies this
issue.

In terms of the aggregator function $\rho_\eta(\cdot)$, modern MIL approaches often employ attention-based aggregation, enabling the model to weight patches according to their inferred relevance and to generate interpretable heatmaps that highlight histologically interesting regions. 
Following this strategy \citep{ilse2018attention}, the attention weight $a_{ij}$ for each refined instance $h_{ij}$ in the bag is defined as 
\begin{equation}\label{eq:mil_attention}
   a_{ij} = \frac{\exp\Big( \mu^\top\text{tanh}\big(\mathbf{V} h_{ij}^\top\big) \Big)}{\sum_{k=1}^{N_i} \exp\Big( \mu^\top\text{tanh}\big(\mathbf{V} h_{ik}^\top\big) \Big)}, 
\end{equation}
where $\eta = \{\mu, \mathbf{V}\}$, with $\mu \in \mathbf{R}^{d_4}$ and $\mathbf{V} \in \mathbb{R}^{d_4 \times d_2}$ are learnable parameters. The bag-level embedding $\mathbf{z}_i$ can be obtained as an attention-weighted sum of refined patch embeddings, i.e.
\begin{equation}\label{eq:slide-level-embedding}
\mathbf{z}_i =  \rho_\eta\left(\{g_\nu(x_{ij})\}_{j=1}^{N_i}\right)=\sum_{j=1}^{N_i} a_{ij}h_{ij}.
\end{equation}
It is important to clarify that the standard MIL attention mechanism \citep{ilse2018attention} is distinct from the self-attention mechanism \citep{vaswani2023attentionneed} that powers modern Large Language Models (LLMs).
The MIL attention in \eqref{eq:mil_attention} is pointwise and static. It derives an importance score $a_{ij}$ by projecting each instance $h_{ij}$ individually onto a fixed, learnable context vector $\eta$ (i.e., $\mu^\top \text{tanh}( \mathbf{V} h_{ij}^T)$). 
This vector $\eta$ acts as a global ``query" that is identical for all slides and all patches, effectively asking: ``How important is this patch to recognize a generic concept of interest (e.g., tumor)?"
In contrast, the attention in Transformers (and Graph Attention Networks, as discussed later) is pairwise and dynamic. 
It computes similarity scores (e.g., dot products) between specific pairs of instances, measuring how relevant instance $\ell$ is to instance $j$ specifically (i.e., $ h_j^\top h_\ell)$).
We will revisit this relational formulation in Section~\ref{sec:gnn} when introducing the Graph Attention Network (GAT) \citep{velickovic2017graph} for local feature refinement.
 
Finally, the slide-level prediction can be written as 
\begin{equation}\label{eq:mil_prediction}
\mathbb{P}(y_i=1)=f_\omega \Big(  \{g_\nu(x_{ij})\}_{j=1}^{N_i} \Big) \quad\text{for $1\leq i\leq n$}.
\end{equation}
Here, $f_\omega = \sigma (\beta_0 + \bm{\beta}^\top\mathbf{z}_i)$ is a logistic classifier that uses slide-level embeddings $\mathbf{z}_i$ in \eqref{eq:slide-level-embedding} and depends on unknown parameters $\omega = (\beta_0, \bm{\beta}, \eta)'$. 
The complete set of learnable parameter is $\theta = (\omega, \nu)'$. In principle, $f_{\omega}(\cdot)$ can be replaced by a deep learning classifier but this was not our focus here.

Intuitively, starting with the bag $B_i$, we proceed to process the individual bag elements $x_{ij}$ through a mapping $g_\nu(\cdot)$ and build a classifier from a bag of refined characterizations $g_\nu(x_{ij})$. 
Slide predictions are obtained from 
\begin{equation*}
\widehat{\mathbb{P}}(y_i=1)=f_{\hat{\omega}} \Big(\{ g_{\hat{\nu}}(x_{ij}) \}_{j=1}^{N_i} \Big),
\end{equation*}
where parameters $\hat{\theta}=(\hat{\omega},\hat{\nu})'$ are estimated by minimizing the cross-entropy loss, i.e. $\hat{\theta}=\arg\min_\theta  L_\theta^{MIL}(\{y_i\}_{i=1}^n)$, where
\begin{align}\label{eq:cross_entropy}
 \mathcal L_\theta^{MIL}(\{y_i\}_{i=1}^n)&= -\frac{1}{n}\sum_{i=1}^n \left[y_i\log \mathbb{P}(y_i=1)\right. \\
 &\quad\quad\quad\left.+(1-y_i)\log\mathbb{P}(y_i=0)\right]. \nonumber
\end{align}
Later, we will augment the loss function $\mathcal{L}_\theta^{MIL}(\{y_i\}_{i=1}^{n})$ with the jigsaw regularizer that will ultimately alter $\hat\theta$ and thereby also predictions $\widehat{\mathbb{P}}(y_i=1)$.

In the following sections, we describe the structure of $g_\nu(\cdot)$ in the context of Graph Neural Networks (GNN) that borrow strength from neighboring patches. 

\section{Spatially Aware Featurization $g_\nu(\cdot)$}\label{sec:spatial_featurization}
To relax the exchangeability implicit in \eqref{eq:mil_attention} and \eqref{eq:slide-level-embedding}, we allow $g_\nu(\cdot)$ to operate on the entire bag $B_i$ instead of just $x_{ij}$.
Specifically, whereas standard MIL computes $h_{ij} = g_\nu(x_{ij})$ solely from the patch itself, our graph-based formulation computes 
$h_{ij} = g_\nu(x_{ij}, \mathcal{N}_{ij}),$
where $\mathcal{N}_{ij}$ denotes the set of ``neighbors" for a patch $x_{ij}$.  

\subsection{Spatial Graph Construction}\label{sec:graph_construction}
We represent  $i^{th}$ whole-slide image  as a graph $\mathcal{G}_i=(V_i,\mathcal{E}_i)$ indexed by vertices $V_i=\{1,...,N_i\}$ that correspond to top-down left-to-right ordering of patches in the bag $B_i$.
The set of edges $\mathcal{E}_i$ is constructed from  $k$ nearest neigbors based on the Euclidean distance between normalized patch centroids $c_{ij} \in [0,1]^2$.
An edge $(j,\ell)\in \mathcal{E}_i$ exists if the centroid $c_{ij}$ is among the $k$ nearest neighbors to $c_{i\ell}$, or conversely, if $c_{i\ell}$ is among the $k$ nearest to $c_{ij}$. 
To explicitly encode proximity, we assign a weight $w_{j\ell}$ to each edge $(j,\ell) \in \mathcal{E}_i$ using a Gaussian kernel
\begin{equation}\label{eq:edge_weight}
w_{j\ell}=\exp\!\left(-\frac{\|c_{ij}-c_{i\ell}\|_2^2}{\sigma^2}\right),
\end{equation}
where $c_{ij}$ denotes the normalized centroid coordinate of patch $j$. 
We set the bandwidth $\sigma$ to the maximum distance observed in the $k$NN graph to penalize edges spanning artificially large distances (e.g., across tissue folds or background).
These weights will be later used in (\ref{eq:gat_attention}).

The hyperparameter $k$ dictates the spatial extent of the local message passing. 
\begin{itemize}
\item If $k$ is too small, the graph becomes overly sparse, effectively reverting the model to the independence assumption where patches are processed in isolation.
\item If $k$ is too large, the attention becomes global (as in Transformer \citep{vaswani2023attentionneed}) rather than local, incurring quadratic computational complexity $\mathcal{O}(N_i^2)$.
\end{itemize}
Therefore, $k$ must be chosen to balance local context integration with computational tractability. 
In our implementation, we empirically set $k=50$.

\subsection{Local Context via Graph Attention}\label{sec:gnn}
Having constructed the graph $\mathcal{G}_i = (V_i, \mathcal{E}_i)$, our instance encoder $g_\nu(\cdot)$ is build around a Graph Attention Network (GAT) \citep{velickovic2017graph}.
Unlike averaging-based graph convolutions (e.g., GCN \citep{kipf2016semi} or GIN \citep{xu2018powerful}) which implicitly smooth representations, GAT selectively emphasizes informative neighbors while attenuating irrelevant ones. 
This adaptive weighting is critical for preserving sharp morphological contrasts across tissue boundaries (e.g., tumor margins), which are often the most clinically relevant regions \citep{adnan2020representation, chen2021whole, pati2022hierarchical}.

The refined patch embedding $h_{ij} = g_\nu(x_{ij}, \mathcal{N}_{ij})$ 
is computed through the following steps:
\begin{enumerate}
    \item \textbf{Linear Projection:} A learnable weight matrix $\mathbf{W}_{G} \in \mathbb{R}^{d_2 \times d_1}$ is applied to every node in the graph. 
    This projects both the central patch $j$ and all its neighbors $\ell \in \mathcal{N}_{ij}$ into a shared latent feature space, enabling comparable feature representation before aggregation.
    \item \textbf{Pairwise Local Attention:}
We compute a pairwise attention coefficient $e^{(i)}_{j\ell}$ that indicates the importance of a neighbor $\ell$ to patch $j$. This is parameterized by a weight vector $\mathbf{v} \in \mathbb{R}^{2d_2}$
\begin{equation}\label{eq:gat-attention-score}
    e^{(i)}_{j\ell} = \text{LeakyReLU}\left(\mathbf{v}^\top [\mathbf{W}_Gx_{ij} \| \mathbf{W}_Gx_{i\ell}]\right),
\end{equation}
here $\|$ denotes concatenation and LeakyReLU is the non-linear activation \citep{velickovic2017graph}.
The formulation in (\ref{eq:gat-attention-score}) effectively measures the relevance of a node $\ell$ to node $j$. 

We then normalize $e^{(i)}_{j\ell}$ via a softmax function over all neighbors $m \in \mathcal{N}_{ij}$  using spatial weights $w_{j\ell}$ \eqref{eq:edge_weight}:
\begin{equation}\label{eq:gat_attention}
   \alpha^{(i)}_{j\ell} = \frac{w_{j\ell}\exp(e^{(i)}_{j\ell})}{\sum\limits_{m \in  \mathcal{N}_{ij} \cup \{j\}} w_{jm} \exp(e^{(i)}_{jm})}.
\end{equation}
The local attention weight $\alpha^{(i)}_{j\ell}$ represents the \emph{local structural attention} determining the relative importance of a neighbor $\ell$ to patch $j$.

Unlike standard GAT implementations driven solely by feature similarity \citep{velickovic2017graph}, our formulation introduces a spatial discounting mechanism through $w_{j\ell}$ defined in \eqref{eq:edge_weight}. 
This explicitly downweights the contribution of remote patches, aligning the message passing with the biological intuition that cellular influence naturally decays with physical distance without being explicitly truncated.

\item \textbf{Context-aware Patch Embedding:} Finally, the refined representation $h_{ij}$ is computed as a weighted sum of the projected neighbor features, followed by a non-linearity $\phi(\cdot)$ (e.g., ELU \citep{clevert2015fast}):
$$h_{ij} = \phi\Big( \sum\limits_{m \in  \mathcal{N}_{ij} \cup \{j\}} \alpha^{(i)}_{jm} \mathbf{W}_{G} x_{im} \Big).$$
\end{enumerate}
Here, the function $g_\nu$ is parametrized by a set of learnable parameters $\nu = \{\mathbf{W}_G, \bm{\alpha}\}$, where $\mathbf{W}_G \in \mathbb{R}^{d_2 \times d_1}$ and $\bm{\alpha} =[\alpha^{(i)}_{12},....,\alpha^{(i)}_{N_i-1,N_i}]' \in \mathbb{R}^{|\mathcal{E}_i|}$. Note that the learnable weights $\mathbf{W}_G$ are distinct from the spatial weight to measure the proximity among patches defined in ($\ref{eq:edge_weight}$).

It is worth contrasting the local structural attention $\alpha^{(i)}_{j\ell}$ in \eqref{eq:gat_attention} with the global pooling attention $a_{ij}$ defined in \eqref{eq:mil_attention}.
Recalling the distinction established in Section~\ref{sec:mil_in_pathology}, standard MIL attention is a pointwise operation that screens isolated instances with fixed and shared parameters $\eta$.
In contrast, the GAT mechanism is pairwise and relational: it computes alignment scores between specific node pairs ($x_{ij}, x_{i\ell}$), conceptually mirroring the ``Query-Key'' self-attention mechanism in Transformers \citep{vaswani2023attentionneed}. 
While GAT asks ``how relevant is neighbor $\ell$ \emph{to me ($j$)}?'', MIL attention asks ``how relevant is patch $j$ \emph{to the diagnosis}?”.

\section{The Jigsaw Regularization}\label{sec:jigsaw-main}
While the refinement $g_\nu(\cdot)$ introduced in the previous section capture local dependencies, they are inherently limited to the neighborhood size defined by $k$. 
Each slide is different and it is conceivable that for certain kinds of tissues, the neighborhood network should be narrower or wider. A prefixed choice of the network may fail to encode global dependencies and information (e.g., ``Are we in the center of the slide or its periphery?'').

To enforce this global awareness, we formulate an auxiliary jigsaw puzzle task. 
We discretize the WSI coordinate space into a coarse grid of $G \times G$ spatial bins, defining the set of target locations as $\mathcal{S}_{\text{bins}} = \{1, \dots, G^2\}$.
To ensure computational efficiency and encourage the model to learn regional tissue semantics rather than precise coordinates, we select a grid resolution such that the number of bins is significantly smaller than the total number of patches (i.e., $G^2 \ll N_i$); in our implementation, we set $G = 10$.
For each patch $j$ in slide $i$ with normalized centroid coordinates $c_{ij} \in [0,1]^2$, we assign a discrete ground-truth label $y^\ast_{ij} \in \mathcal{S}_{\text{bins}}$ corresponding to the spatial grid cell containing that patch.

We augment the model with an auxiliary classifier head $f^{\text{aux}}_\psi: \mathbb{R}^{d_2} \to \mathbb{R}^{G^2}$, parameterized by $\psi$. 
This head predicts the absolute spatial grid bin of a patch based solely on its GNN-refined embedding $h_{ij}$. The probability that the $j^{th}$ patch belongs to the $k^{th}$ grid bin is given by the softmax distribution:
\begin{equation}\label{eq:jigsaw_prediction}
\mathbb{P}(y_{ij}^\ast = k \mid h_{ij}) = \frac{\exp(f^{\text{aux}}_\psi(h_{ij})_k)}{\sum_{c=1}^{G^2} \exp(f^{\text{aux}}_\psi(h_{ij})_c)},
\end{equation}
where $f^{\text{aux}}_\psi(h_{ij})_k$ denotes the logit for the $k^{th}$ spatial bin. 
Now, the complete learnable parameter set becomes $\theta =(\omega, \nu, \psi)'$.

\begin{figure}[!t]
    \centering
    \includegraphics[width=1\linewidth]{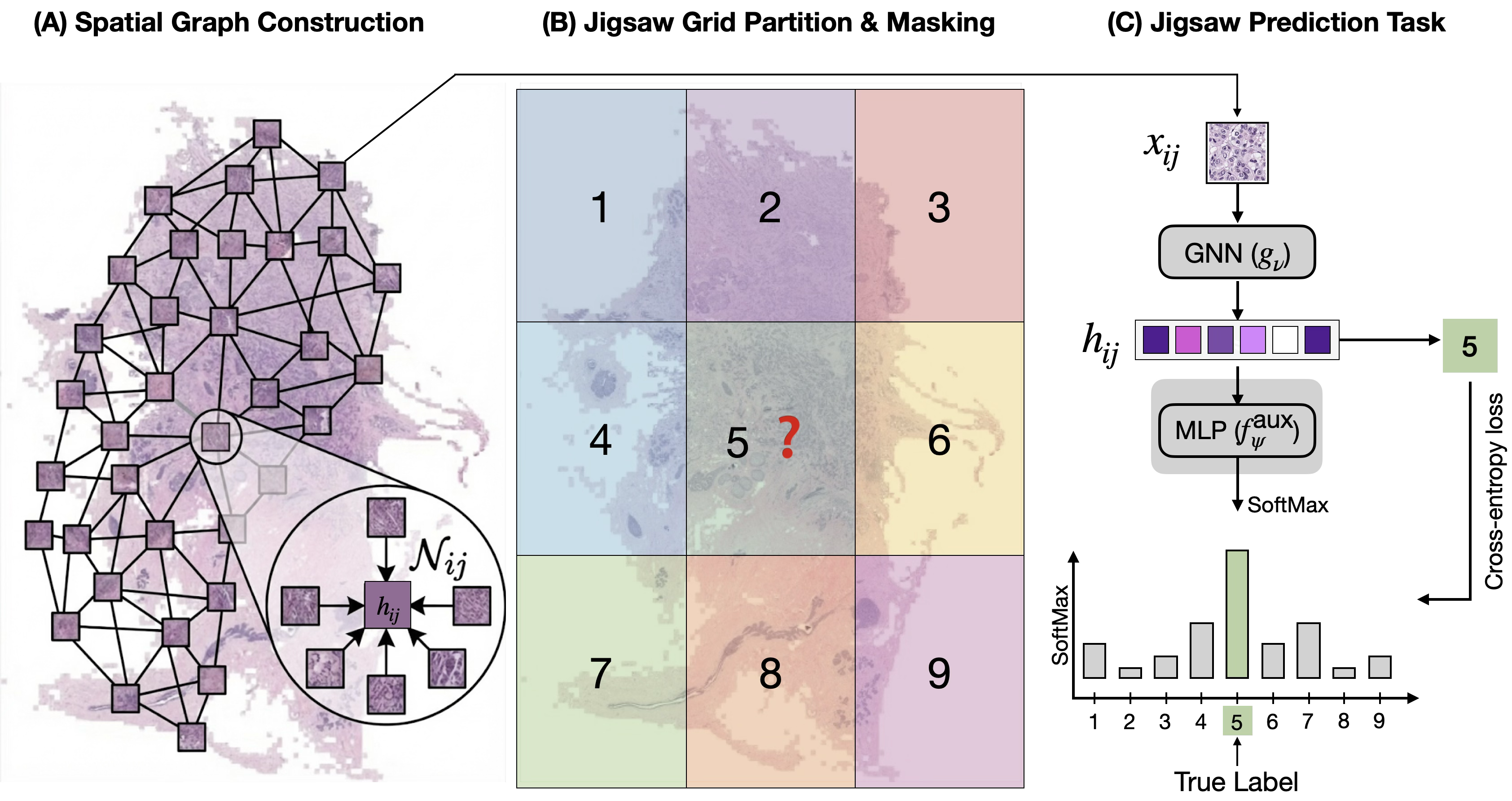}
    \caption{Jigsaw auxiliary task pipeline integrated into graph-MIL.}
    \label{fig:jigsaw-implmentation}
\end{figure}

The jigsaw penalty for the $i^{th}$ slide is defined as the cross-entropy between the predicted location and the true spatial bin.
To compute this, we employ a stochastic masking strategy where we select a random subset of patches $\mathcal{M}_i \subset \{1, \dots, N_i\}$ for supervision. The loss for the $i^{th}$ bag is then
\begin{equation*}
\mathcal{L}^{\text{jigsaw}}(i) =
- \frac{1}{|\mathcal{M}_i|}\sum_{j \in \mathcal{M}_i} \sum_{k=1}^{G^2} \mathbb{I}(y_{ij}^\ast = k) \log\mathbb{P}(y_{ij}^\ast = k \mid h_{ij}).
\end{equation*}
The total regularization term is the average over $n$ slides:
\begin{equation*}
\mathcal{L}^{\text{jigsaw}} = \frac{1}{n} \sum_{i=1}^n \mathcal{L}^{\text{jigsaw}}(i).
\end{equation*}
The set $\mathcal{M}_i$ is a random subsample of patches drawn uniformly from each bag, rather than the full set $\{1, ..., N_i\}$. 
In our implementation, we retain approximately 90\% of patches for jigsaw supervision in each training step (i.e., $|\mathcal{M}_i| \approx 0.9 N_i$), resampling $\mathcal{M}_i$ independently at each iteration. 
This stochastic subsampling serves two purposes. 
First, it acts as a regularizer on the auxiliary task, by varying which patches are supervised across epochs, we prevent the jigsaw head from trivially memorizing fixed patch-to-bin assignments and instead force the shared encoder $g_\nu$ to learn generalizable spatial representations. 
Second, it reduces the computational cost of the auxiliary loss, which scales linearly in $|\mathcal{M}_i|$. 
Crucially, the GNN encoder still processes all patches and all edges during forward propagation and only the supervision signal from the jigsaw head is sparsified, not the message-passing computation.
Figure~\ref{fig:jigsaw-implmentation} illustrates the construction of the spatial bins $\mathcal{S}_\text{bins}$ used for the jigsaw auxiliary task.

While we do not explicitly shuffle patches as in classical jigsaw games, the challenge remains mathematically equivalent: the model must be aware of the patch positions.
Intuitively, minimizing   $\mathcal{L}^\text{jigsaw}$ forces the encoder parameters $\nu$ to embed spatial coordinates directly into the feature $h_{ij}$. 
The final training objective combines the clinical goal with the spatial constraint:
\begin{equation*}
    \hat{\theta} = \arg\min_{\theta} \left( \mathcal{L}^{MIL}_{\nu, \omega} + \lambda \mathcal{L}^{\text{jigsaw}}_{\nu, \psi} \right).
\end{equation*}
Here, $\lambda$ is a regularization weight that balances the trade-off between diagnostic accuracy and spatial reasoning.

\section{Spatially-Aware Slide Embeddings}
The final component of our approach is the aggregation of patch features into a slide-level representation.
Standard MIL architecture enforces a bag $B_i = \{x_{i1}, \dots, x_{i N_i}\}$ is composed of \emph{exchangeable} instances, necessitating that the bag-level probability $\mathbb{P}(y_i=1 \mid B_i)$ be a permutation-invariant function of $B_i$ \citep{ilse2018attention}. 
According to Theorem 1 in \citet{ilse2018attention}, any such permutation-invariant function can be decomposed as 
\begin{equation*}
    \mathbb{P}(y_i = 1 \mid B_i)  = f \Big( \rho \big( g(\{x_{ij}\}_{j=1}^{N_i}\big) \Big),
\end{equation*}
where $g$ is a continuous instance-wise transformation, $\rho$ is a symmetric aggregation operator (e.g., sum or max), and $f$ is a continuous mapping to the bag score. 
This decomposition mathematically constrains the model to treat the WSI as an unordered bag.

In contrast, our framework intentionally violates this exchangeability assumption by inducing  spatial couplings. 
We replace the instance-wise mapping $g$ with a context-dependent function $g_\nu$ (as defined in Section~\ref{sec:gnn}) that conditions the embedding $h_{ij}$ on its local neighborhood $\mathcal{N}_{ij}$ and  its global position $y_{ij}^\ast \in S_{\text{bins}}$:
\begin{equation*}
    h_{ij} = g_\nu(x_{ij} \mid \mathcal{N}_{ij}, y_{ij}^\ast).
\end{equation*}

\begin{figure}
    \centering
    \includegraphics[width=0.8\linewidth]{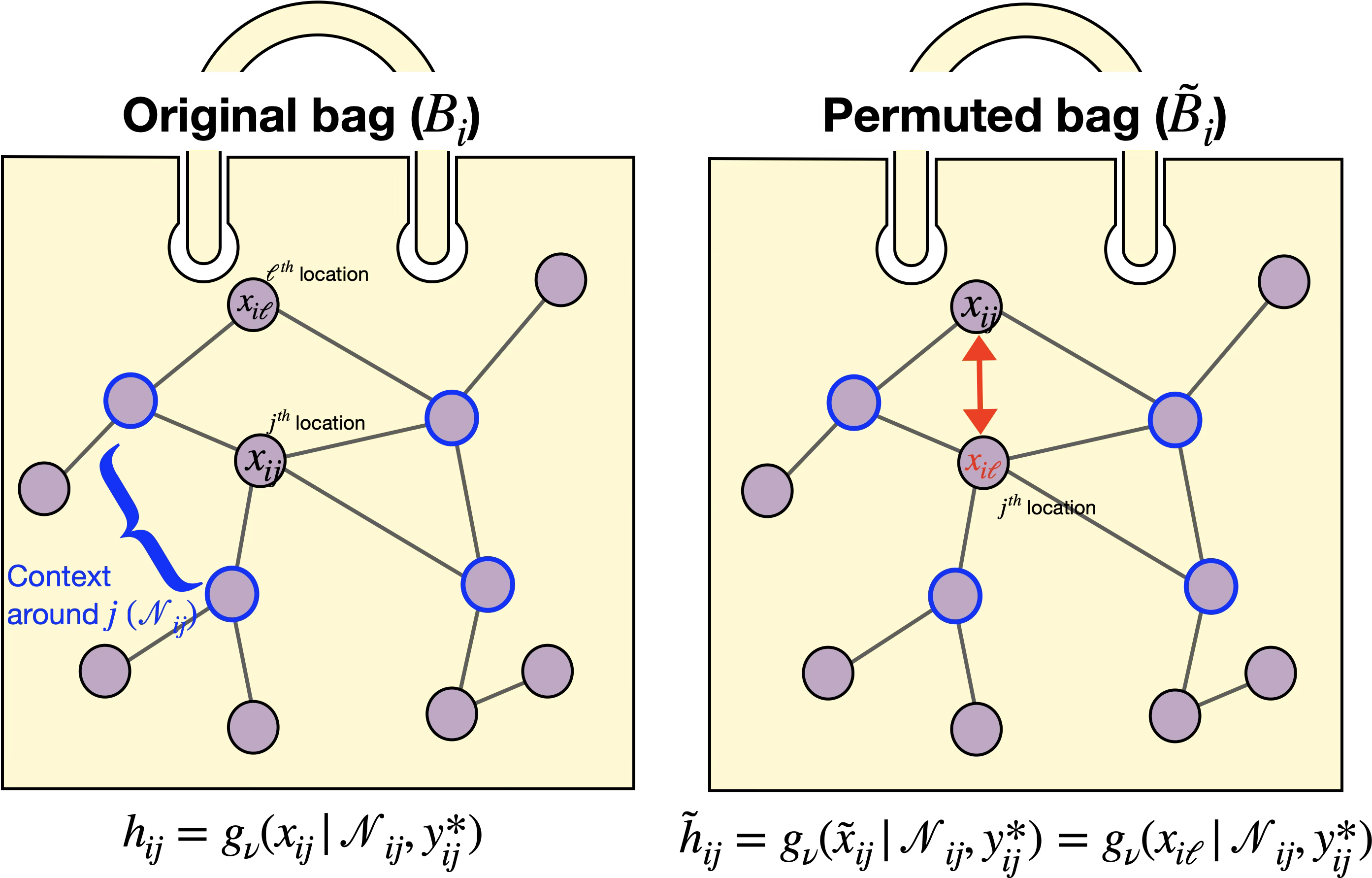}
    \caption{Spatial permutation of patches alters their embeddings $\tilde{h}_{ij}$ under our spatially-aware framework.}
    \label{fig:spatial-permutation}
\end{figure}

Consider a spatial permutation where the patch content originally at location $\ell$   is moved to location $j$. 
Let $\tilde{x}_{ij}$ denote the input at location $j$ in this permuted bag, such that $\tilde{x}_{ij} = x_{i\ell}$.
In a standard MIL model, the feature refinement would remain identical, $g(x_{i\ell}) = g(\tilde{x}_{ij})$.
However, in our framework, the feature generation depends on the spatial context $(\mathcal{N}_{ij}, y_{ij}^\ast)$ as well as its own feature  $x_{ij}$.
At location $j$, the encoder $g_\nu(\cdot)$ still observes the fixed spatial content of location $j$ (neighbors $\mathcal{N}_{ij}$ and absolute location $y_{ij}^\ast$), but it now receives the feature from $\ell^{th}$-patch:
\begin{equation}
\tilde{h}_{ij} = g_\nu(x_{i\ell} \mid \mathcal{N}_{ij}, y_{ij}^\ast).
\end{equation}
Because the $\ell^{th}$-patch is now embedded using the mismatched spatial context at location $j$, the resulting feature value changes (i.e., $\tilde{h}_{ij} \neq h_{i\ell}$).
See Figure~\ref{fig:spatial-permutation} for an illustration.
This renders the set of embeddings for the permuted bag $\tilde{H}_i = \{\tilde{h}_{i1}, \dots, \tilde{h}_{iN_i}\}$ is numerically distinct from the original bag $H_i$ (i.e., $\tilde{H}_i \neq H_i$).

Consequently, although the final aggregation function $\rho_\eta(\cdot)$ (e.g., Attention-MIL pooling) remains a symmetric operator, the input set passed to it has been fundamentally transformed by the spatial permutation:
\begin{equation}
\tilde{\mathbf{z}}_i = \rho_\eta(\tilde{H}_i) \neq \rho_\eta(H_i) = \mathbf{z}_i.
\end{equation}
This in turn establishes the overall scoring function $\mathbb{P}(y_i = 1 \mid B_i)$ \emph{non-exchangeable} with respect to the spatial arrangement of patches.

It is important to note that our final aggregation operator $\rho_\eta(\cdot)$ retains the identical functional form as the standard attention pooling defined in \eqref{eq:mil_attention}, remaining a mathematically symmetric (permutation-invariant) weighted sum. 
However, we break the global exchangeability of the slide scoring function $\mathbb{P}(y_i \mid B_i)$ by breaking the independence of the input embeddings $H_i$. 
By coupling feature generation to spatial coordinates via $g_\nu(\cdot)$, we ensure that the aggregation mechanism becomes \emph{spatially aware}, allowing the model to distinguish distinct tissue architectures that standard MIL would perceive as identical.

\section{Parameter Tuning}\label{sec:bayesian-calibration}
The model is trained end-to-end by minimizing the joint objective:
\begin{equation}\label{eq:total_loss}
\mathcal{L}^{\text{tot}}_\theta(\lambda) = \mathcal{L}^{MIL}_{\nu, \omega} + \lambda \mathcal{L}^{\text{jigsaw}}_{\nu, \psi}.
\end{equation}
In multitask learning, achieving balance among tasks  is the key to   successful training \citep{kendall2018multi}.
We employ an EM-inspired approach by treating $\lambda$ as a random variable with a conjugate Gamma prior, $\lambda \sim \text{Gamma}(\alpha, \beta)$. 
For a fixed set of parameters $\theta$, \eqref{eq:total_loss} can be viewed as a log-likelihood in $\lambda$ under  an exponential ``pseudo-likelihood" that is proportional to $ \exp(-\lambda \mathcal{L}^{\text{jigsaw}}_{\nu, \psi})$.
Treating $\lambda$ as the missing data in the EM framework, one could replace $\lambda$ with a conditional expectation given the current set of parameters.
The posterior expectation with respect to $\lambda$ thus offers an iterative update of the tuning parameter according to
\begin{equation}\label{eq:lambda_update}
\mathbb{E}[\lambda \mid \mathcal{L}^{\text{jigsaw}}_{\nu, \psi}] = \frac{\alpha}{\beta + \mathcal{L}^{\text{jigsaw}}_{\nu, \psi}}.
\end{equation}
The regularization strength $\lambda$ is small when the jigsaw task is difficult (early training), preventing auxiliary noise from dominating the gradient, and increases as the model learns to resolve spatial structures.

Since $\mathcal{L}^{\text{jigsaw}}_{\nu, \psi}$ could exhibit high variance across mini-batches due to sampling stochasticity, updating $\lambda$ at every iteration can destabilize the optimization trajectory.
To mitigate this, we update $\lambda$ every $n_\lambda$ steps (e.g., once per $n_\lambda$ epoch). 
This ensures that the posterior update is conditioned on a robust estimate of the loss (averaged over the epoch) rather than a noisy single-batch estimate.
See Algorithm~\ref{alg:graphmil-jigsaw} for the full pipeline of our proposed approach.

\begin{algorithm}[!t]
\caption{Graph-MIL with Jigsaw Regularization}
\label{alg:graphmil-jigsaw}
\begin{algorithmic}[1]
\REQUIRE 
Whole-slide images (WSIs) $i \in [n]$, and corresponding path foundational embeddings $\{x_{ij}\}_{j=1}^{N_i}$, learning rate $\gamma$ 

\STATE \textbf{Graph construction (Section~\ref{sec:graph_construction}):}
\STATE \hspace{1em} Compute edges $\mathcal{E}_i$ based on spatial proximity.
\STATE \hspace{1em} Define graph $\mathcal{G}_i = (V_i, \mathcal{E}_i)$ where each node  $j$ has a patch embedding $x_{ij} \in \mathbb{R}^{d_1}$ and neighbors $\mathcal N_{ij}$.
\vspace{0.3em}
\STATE  \textbf{Shared GNN encoder $g_\nu(\cdot)$:}
\FOR{each node $j \in V_i$}
    \STATE Refine embedding: $h_{ij} \leftarrow g_\nu(x_{ij}, \mathcal{N}_{ij})$ 
\ENDFOR

\vspace{0.3em}
\STATE  \textbf{Dual task heads:}
\STATE \hspace{1em} \textit{(a) MIL Head:} \\
Aggregate $\{h_{ij}\}$ via attention pooling (\ref{eq:mil_attention}) to predict slide label:
 $\hat{y}_i  = f_\omega\big(\{h_{ij}\}_{j=1}^{N_i}\big)$ as in  (\ref{eq:mil_prediction}).

\STATE \hspace{1em} \textit{(b) Jigsaw Head:}\\
 Predict  bin label $\hat{y}^\ast_{ij} = f^{\text{aux}}_\psi (h_{ij})$ as in (\ref{eq:jigsaw_prediction}).

\vspace{0.3em}
\STATE  \textbf{Compute losses (\ref{eq:total_loss}):}
\STATE \hspace{1em} $\mathcal{L}_\theta = \sum_{i} \mathcal{L}^{\text{MIL}}_{\nu, \eta}(\hat{y}_i, y_i) + \lambda \sum_{i} \mathcal{L}^{\text{jigsaw}}_{\nu, \psi} (\hat{y}^\ast_{ij}, y^\ast_{ij})$
\vspace{0.3em}
\STATE \textbf{Parameter update:}
\STATE \hspace{1em} Update shared encoder $\nu$ and task heads ($\omega, \psi$):
\STATE \hspace{2em} $\{\nu, \omega, \psi\} \leftarrow \{\nu, \omega, \psi\} - \gamma \nabla_{\{\nu, \omega, \psi\}} \mathcal{L}_\theta$
\STATE \hspace{1em} Update $\lambda$ every $n_\lambda$ steps via EM-like strategy (\ref{eq:lambda_update}).

\RETURN labels $\{\hat{y}_i\}_{i=1}^n$ and attentions $\{\{\hat{a}_{ij}\}_{j=1}^{N_i}\}_{i=1}^n$
\end{algorithmic}
\end{algorithm}

\section{Experiment}\label{sec:experiment}
To validate the proposed Spatially-Aware Graph-MIL framework, we conduct experiments designed to address two complementary questions: (i) whether the method generalizes across different cancer types and diagnostic tasks, and (ii) whether the observed performance gains depend on a specific feature representation or hold across commonly used histopathology foundation models.

\begin{table}
\caption{Performance across Cancer Types on small TCGA benchmarks. 3-fold cross-validated ROC-AUC is reported. }
\centering
\resizebox{0.9\linewidth}{!}{%
    \begin{tabular}{lccc}
        \toprule
        \textbf{Method} & BRCA & HNSC & COAD \\
        \midrule
        ABMIL  & 0.884 & 0.809 & 0.822 \\
        ABMIL+Jigsaw & 0.880 & 0.820 & 0.806 \\
        Graph-MIL & 0.900 & 0.830 & \textbf{0.901} \\
        Graph-ABMIL & 0.932 & 0.810 & 0.841 \\
        Graph-ABMIL + Jigsaw & \textbf{0.952} & \textbf{0.833} & 0.885 \\
        \bottomrule
    \end{tabular}
}
\label{tab:auc_comparison_across_cancer_types}
\end{table}

\begin{figure}[!t]
    \centering
    \includegraphics[width=\linewidth]{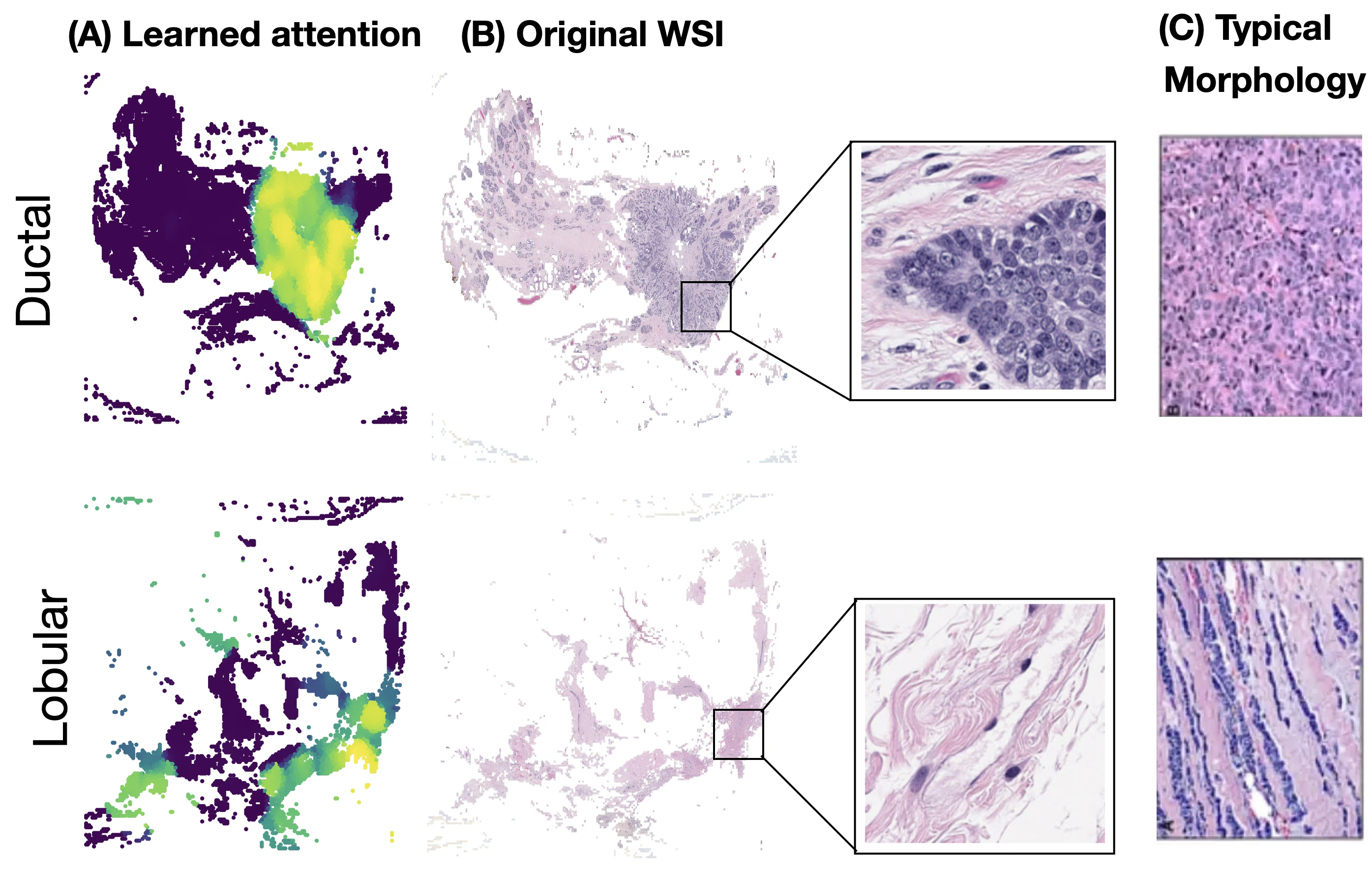}
    \caption{Visualization of attention map. Here, the attention weight ($a_{ij}$) is computed following (\ref{eq:mil_attention}). Each point corresponds to each patch. The trained model is evaluated on the test data points. It is known that ductal and lobular breast cancer has visually different morphology \footnote{https://lobularbreastcancer.org/its-not-just-breast-cancer-its-lobular-breast-cancer/}, which serves as a reference point in the third column. }
\label{fig:attention_map_brca}
\end{figure}

\subsection{Datasets and Experimental Setup}

We evaluate our method on three benchmarks from The Cancer Genome Atlas (TCGA) \citep{weinstein2013cancer}, spanning distinct tissue types and classification objectives: (a) Breast Cancer (BRCA) (ductal vs. lobular subtyping, $n=100$), (b) Head \& Neck (HNSC) (HPV status, $n=101$), and (c) Colon (COAD) (MSI status, $n=78$). 
For these experiments, we utilize UNI embeddings \citep{chen2024towards} to fix the feature representation and isolate the effects of model architecture.
To assess the contribution of each component, we compare our approach against standard ABMIL \citep{ilse2018attention} and controlled ablations. 
While various graph-based MIL methods exist \citep{lu2020capturing, zhao2020predicting, ma2024gcn}, we restrict our analysis to our specific spatial backbone to explicitly decouple the effects of local topological modeling from auxiliary regularization. 
The evaluation includes (a) ABMIL \citep{ilse2018attention}; (b) ABMIL + Jigsaw (regularization only); (c) Graph-MIL (Graph with average pooling, i.e. $\rho_\eta(\cdot)$ is simaly an average) (d) Graph-ABMIL (No jigsaw, construction in Section~\ref{sec:spatial_featurization}); and (e) Graph-MIL + Jigsaw (Our proposal). 
Results are reported in Tables~\ref{tab:auc_comparison_across_cancer_types} and \ref{tab:auc_comparison_across_feature_types}.

Finally, we assess the method's robustness to different feature representations using the larger TCGA Breast Invasive Carcinoma (BRCA) cohort ($n=1,133$). 
In this setting, we fix the downstream architecture and vary the input embeddings across four widely adopted foundation models: \textit{GigaPath} \citep{xu2024whole}, \textit{RetCCL} \citep{wang2022transformer}, \textit{UNI} \citep{chen2024towards}, and \textit{Virchow2} \citep{vorontsov2024foundation}. 
A summary of embedding characteristics is provided in Appendix Table~\ref{tab:tcga_brca_data_summary}.

\subsection{Analysis of Results}

\paragraph{Impact of Graph Structure:} The introduction of the graph structure (Graph-ABMIL vs. ABMIL) yields a clear performance improvement in the Breast Cancer (BRCA) dataset, increasing the AUC from 0.884 to 0.932 (+6.6\%)
This gain suggests that for breast cancer, the local neighborhood (e.g., tubule formation, specific cell-cell interactions) is highly diagnostic. 
The $k$NN-GAT structure successfully captures these local dependencies, which are missed by the independent processing of ABMIL.

\paragraph{Impact of Jigsaw Regularization:} Adding the Jigsaw regularization consistently improves performance across all datasets. 
In HNSC (Head \& Neck) dataset, the pure Graph-ABMIL performed comparably to simple ABMIL (0.810 vs 0.809), possibly due to dispersed tumor cells that do not form coherent local graphs. 
However, adding the Jigsaw constraint significantly boosted the performance to 0.833 (+2.8\%).
This suggests that the global structural prior helps regularize the graph learning against noisy edges, making the model more robust. 

\paragraph{Impact of Attention under Graph Modeling:} 
Comparing Graph-MIL and Graph-ABMIL isolates the role of attention at different stages of the pipeline. 
Graph-MIL aggregates graph-refined node embeddings using simple mean pooling, whereas Graph-ABMIL applies an additional attention-based MIL pooling layer \citep{ilse2018attention} after graph message passing. 
As shown in Tables~\ref{tab:auc_comparison_across_cancer_types} and~\ref{tab:auc_comparison_across_feature_types}, the performance difference between these two variants is minimal.
This indicates that once local spatial context is explicitly modeled via the GAT encoder, much of the discriminative signal is already encoded in the node representations themselves. 
Still, Graph-ABMIL does not degrade performance, suggesting that attention remains compatible with graph-based representations. 

In this setting, the primary benefit of attention is an interpretable mechanism for visualizing which spatially contextualized patches contribute most to the slide-level prediction, enabling intuitive localization of diagnostically relevant regions without being the primary driver of performance gains. 
Figure~\ref{fig:attention_map_brca} demonstrates that our method not only improves predictive performance but also identifies patch regions that are most influential for the final prediction.

\begin{table}
\caption{Performance of TCGA-BRCA dataset across different foundation embeddings. 10-fold cross-validated ROC-AUC is reported.}
\label{tab:auc_comparison_across_feature_types}
\centering
\resizebox{0.49\textwidth}{!}{%
    \begin{tabular}{lcccc}
        \toprule
        \textbf{Method} & \textbf{GigaPath} & \textbf{RetCCL} & \textbf{UNI} & \textbf{Virchow2}\\
        \midrule
        ABMIL & 0.941& 0.919& 0.941&  0.936\\
        Graph-MIL& 0.949 & 0.911 & \textbf{0.954} & 0.948 \\
        Graph-ABMIL  & 0.948 & 0.921 & \textbf{0.954} & 0.936 \\
        Graph-ABMIL + Jigsaw  & \textbf{0.952} & \textbf{0.921} & \textbf{0.954} & \textbf{0.950} \\
        \bottomrule
    \end{tabular}
}
\end{table}

\paragraph{Sensitivity to the Choice of Foundation Model:} 
Table~\ref{tab:auc_comparison_across_feature_types} shows that the proposed Graph-ABMIL + Jigsaw exhibits strong robustness across different of foundation models \citep{xu2024whole, wang2022transformer,chen2024towards,vorontsov2024foundation}, despite substantial differences in embedding dimensionality, patch resolution, and pretraining objectives. 
Across all embeddings, Graph-MIL variants consistently achieve high test ROC-AUC, with only minor variation in absolute performance. 
This suggests that the gains of our method are not driven by properties of a specific feature extractor, but instead arise from the spatial inductive bias imposed by graph-based message passing and auxiliary regularization. 
Notably, embeddings with fewer but semantically richer patches (e.g., RetCCL) and those with higher patch density (e.g., UNI, Virchow2) both benefit similarly from the proposed architecture.

\paragraph{Effectiveness of Bayesian Update on Regularization Strength Parameter $\lambda$:}

In Table~\ref{tab:auc_comparison_lambda_ablation}, we analyzed the effect of the regularization strength of the auxiliary jigsaw task.
\begin{itemize}
    \item \textbf{Fixed $\lambda$:} We performed a grid search over $\lambda \in \{0.01, 0.1, 0.5, 1.0, 2.0, 5.0\}$. Results indicated that fixed weighting is highly sensitive to the dataset. 
    For BRCA, $\lambda=0.5$ achieved the highest AUC of $0.952$, but for COAD the same setting yielded the lowest performance ($0.814$) with high variance ($\pm 0.139$), indicating instability. 
    Conversely, COAD favored either very small ($\lambda=0.01$, AUC $0.848$) or very large ($\lambda=5.0$, AUC $0.860$) values, while these same extremes degraded BRCA performance. 
    HNSC was comparatively robust across settings but peaked at the smallest value ($\lambda=0.01$, AUC $0.843$). 
    This inconsistency across cancer types underscores the difficulty of selecting a single fixed $\lambda$ in practice.
    \item\textbf{EM Update:} Our proposed dynamic weighting scheme eliminates the need for manual tuning, achieving AUC of $0.857$, $0.828$, and $0.809$ on BRCA, HNSC, and COAD respectively without any dataset-specific hyperparameter selection. 
    On COAD, where fixed $\lambda$ exhibited the greatest instability (variance up to $\pm 0.139$), the EM update maintains $\pm 0.092$, which is a meaningful reduction given the small cohort size, while remaining competitive with mid-range fixed settings. 
\end{itemize}

\begin{table}[!t]
\caption{Performance across Cancer Types on small TCGA benchmarks across different regularization strength $\lambda$ compared to the Bayesian updating scheme introduced in Section~\ref{sec:bayesian-calibration}. 3-fold cross-validated test ROC-AUC is reported.}
\label{tab:auc_comparison_lambda_ablation}
\centering
\resizebox{\linewidth}{!}{%
    \begin{tabular}{lccc}
        \toprule
        \textbf{$\lambda$} & \textbf{BRCA} & \textbf{HNSC} & \textbf{COAD} \\
        \midrule
        0.01 & 0.929 $\pm$ 0.020 & 0.843 $\pm$ 0.042 & 0.848 $\pm$ 0.099\\
        0.1 &  0.941 $\pm$ 0.031  &0.832 $\pm$ 0.086 & 0.837 $\pm$ 0.043\\
        0.5 &  0.952 $\pm$ 0.046 &  0.833 $\pm$ 0.06 & 0.814 $\pm$ 0.139\\
        1.0 &0.929 $\pm$ 0.026 & 0.832 $\pm$ 0.060 & 0.846 $\pm$ 0.107\\
        2.0 &0.910 $\pm$ 0.096 & 0.822 $\pm$ 0.083 & 0.817 $\pm$ 0.097\\
        5.0 & 0.879 $\pm$ 0.108 & 0.820 $\pm$ 0.060 & 0.860 $\pm$ 0.053\\
        \midrule
        EM Update & 0.857 ± 0.083 & 0.828 ± 0.043 & 0.809 ± 0.092 \\
        \bottomrule
    \end{tabular}
}
\end{table}

\section{Concluding Remarks}
The results presented in this paper underscore the necessity of moving beyond the ``bag of instances" assumption in digital pathology. 
By treating WSIs as structured graphs and enforcing global spatial consistency via jigsaw puzzles, we align the computational model with the biological reality of tissue architecture.
Our model outperforms the state-of-the-art attention-based MIL by a non-negligible amount which could have significant implications in the practical deployment of digital pathology in everyday diagnostic practice.

\section*{Acknowledgements}
This work was completed in part with resources provided by the University of Chicago’s Center for Research Informatics (CRI).
The CRI is funded by the Biological Sciences Division at the University of Chicago with additional funding provided by the Institute for Translational Medicine, CTSA grant number 2U54TR002389-06 from the National Institutes of Health.

This research was supported by the NSF (DMS: 2515709).




\newpage
\bibliography{main}
\bibliographystyle{icml2026/icml2026}

\newpage
\appendix
\onecolumn

\section{Related Works}\label{sec:related_works}
We review prior work most closely related to our approach, with an emphasis on clarifying both similarities and key distinctions. 
Rather than providing an exhaustive survey, we focus on methods that share architectural components or modeling goals with our framework and highlight how our design departs from these lines of work.

\paragraph{Relation to prior GNN applications on MIL.}
Early extensions of MIL to dependent bags, such as MIGraph and miGraph \citep{zhou2009multi}, represent each bag as a graph and compare bags via graph kernels. 
These approaches predate modern GNNs and end-to-end patch encoders, and do not perform spatial message passing over patch graphs within a slide. 

More recent work combines GNN encoders with MIL pooling, including GCN+ABMIL hybrids \citep{ma2024gcn}. 
A large body of work builds spatially-aware models by constructing spatial or feature-based graphs over WSI patches and applying local message passing \citep{li2018graph, lu2020capturing, mesquita2020rethinking, zhao2020predicting, chen2021whole}. 
These methods focus on local aggregation and often rely on feature similarity or spectral filters to encode structure. 
Other approaches inject spatial context through external priors such as segmentation masks (e.g., SAM-MIL) \citep{fang2024sam}. 

While architecturally related, such methods primarily rely only on local aggregation and typically lack an explicit mechanism for enforcing global spatial structure. 
In contrast, we operate on a strictly spatial instance graph and augment local message passing with a global jigsaw regularizer.

\paragraph{Relation to prior Bayesian MIL}
Bayesian formulations of MIL have been proposed to improve robustness and interpretability, for example through sparsity-inducing priors on instance relevance \citep{chen2025sib}.
Related work on uncertainty-aware learning often targets predictive uncertainty (e.g., MC dropout or entropy-based measures) or relies on manually tuned multi-task weights \citep{gheshlaghi2025uncertainty}. 
Our focus is to view MIL and jigsaw as a multi-task problem with weights as latent variables. 
By introducing a latent regularization weight with a Gamma prior (optionally per slide), we obtain a data-adaptive calibration between objectives, replacing heuristic $\lambda$ tuning with an estimator grounded in likelihood principles.

\subsection{Relation to Prior Jigsaw Literature}
The jigsaw puzzle task is defined as the recovery of an image's original configuration from shuffled patches.
Solving jigsaw puzzle itself has long been a pivotal problem in computer vision \citep{freeman2006apictorial, paumard2020deepzzle, song2023siamese, heck2025solving}.

We focus on Jigsaw task in different contexts (self-supervised objective, ViT, and MIL) and connect their approaches to ours. 

\paragraph{Jigsaw as Self-supervised Task.}
\citet{noroozi2016unsupervised} first demonstrated that training Convolutional Neural Networks (CNNs) to reorder these patches forces the model to learn spatial dependencies, thereby benefiting downstream visual recognition tasks. 
Building upon this work, many subsequent methods have leveraged the jigsaw strategy to learn rich feature representations in a \emph{self-supervised manner} \citep{carlucci2019domain}. 

This paradigm serves as a visual analogue to the ``next word prediction" task ubiquitous in Large Language Models (LLMs); just as predicting the continuation of a sentence forces a model to learn linguistic syntax and semantics, solving visual puzzles compels the network to master the geometric ``grammar" and structural coherence of images. 

\paragraph{Jigsaw in ViT}
Recent literature has further expanded this paradigm into modern architectures such as Vision Transformers (ViT). \citet{chen2023jigsaw} proposed Jigsaw-ViT, exploring the jigsaw puzzle problem as a self-supervised auxiliary loss within the standard ViT framework for image classification. 
Let an image $I$ be decomposed into a sequence of $N$ patches, denoted as $\mathcal{P} = \{p_1, p_2, \dots, p_N\}$. 
During the pretext task (jigsaw puzzle solving), a permutation $\pi \in S_N$ is applied to this sequence, yielding a shuffled input $X_{shuffled} = [p_{\pi(1)}, p_{\pi(2)}, \dots, p_{\pi(N)}]$. 
The model must predict the original absolute position index $y_i \in \{1, \dots, N\}$ for each patch in the shuffled sequence.
The optimization objective is the standard Cross-Entropy loss averaged over all patches (\ref{eq:cross_entropy}). 
Crucially, Jigsaw-ViT removes the positional embeddings from the input of this specific branch ($x_{jigsaw} = x_{patch}$ rather than $x_{patch} + x_{pos}$), forcing the model to infer $y_i$ solely from the semantic content and boundary continuity of the patches $p_i$, where $x_{pos}$ denotes the explicit positional encoding.
Empirical results from this study confirmed that adding the jigsaw branch significantly improved the model's generalization on ImageNet and its robustness against noisy labels and adversarial attacks.

\paragraph{Jigsaw in MIL}
In the context of Multiple Instance Learning (MIL) for histopathology, the objective shifts from finding explicit positions to maintaining feature consistency \citep{chen2025cracking}. 
The authors formulate this using a Siamese network \citep{chicco2021siamese} that processes a bag of instances $X_i \in \mathbb{R}^{N_i \times d_i}$ (where $N_i$ is the number of patches of $i^{th}$ slide).
Let $S$ be a random permutation matrix. 
The ``Jigsaw" constraint is satisfied if the feature extractor $g(\cdot)$ is equivariant to shuffling.
That is, the features of the shuffled input $g(S \cdot X)$ should be identical to the shuffled features of the original input $S \cdot g(X)$. 
This is minimized via the Frobenius norm of the difference:
\begin{equation*}\label{eq:jigsaw-siamese}
    \mathcal{L}_{SER} = \left\| g(S \cdot X) - S \cdot g(X) \right\|_F^2
\end{equation*}
Mathematically, this regularizes the network by penalizing any feature deviations that arise solely from the permutation of input tiles. 
By minimizing this ``transport cost," the model implicitly learns the spatial correlations between patches, as only a model that understands the global tissue structure can produce features stable enough to satisfy this equivalence equation.

Unlike \citet{chen2025cracking}, who focus on feature stability under permutation, our goal is explicit positional localization. 
We adopt a strategy closer to Jigsaw-ViT but adapted for Graph MIL, where jigsaw task is an auxiliary task to complement the positional encoding \citep{chen2023jigsaw}.
We posit that for a node embedding $h_{ij}$ to be truly ``context-aware," it must contain sufficient information to identify its absolute location in the tissue coordinate space.

\section{Alternative Implementation of Jigsaw Auxiliary Task}\label{sec:jigsaw-alternative}

In Section~\ref{sec:jigsaw-main}, we proposed the ``positional jigsaw'' strategy, which enforces spatial awareness by requiring the model to triangulate its \emph{absolute} location on a grid.
This approach is both scalable, as the complexity is controlled by the grid size $G$, and non-trivial, since the GNN’s receptive field is limited to $T$-hop neighbors (typically covering only a fraction of the slide).
However, one might argue that this formulation deviates from the classical definition of a ``jigsaw" puzzle, which inherently involves the reordering of permuted pieces rather than static localization.

To address this and provide a formulation more aligned with the intuitive concept of ``fitting pieces together," we propose an alternative auxiliary task to test the neighborhood consistency.
Unlike the positional approach which asks \textit{``Where am I?"}, this formulation asks \textit{``Do I belong here?"}
By explicitly shuffling patch features to create ``imposter" nodes, we force the shared encoder to learn a discriminative representation of tissue coherence, ensuring that a patch's embedding is strictly conditioned on its compatibility with its local neighborhood. 

The core hypothesis is that a valid tissue representation $h_{ij}$ arises only when the patch content $x_{ij}$ matches its surrounding micro-environment $\mathcal{N}_{ij}$. If a patch feature is transplanted into a mismatched location (e.g., a ``fat" patch placed inside a ``tumor" region), the encoder should recognize this as an anomaly.

This formulation aligns with self-supervised graph learning (similar to Deep Graph Infomax \citep{veličković2018deep}), adapted for histopathology.
We define a Discriminator $D_\psi: \mathbb{R}^{d_2} \to [0,1]$ that estimates the probability that a node embedding represents a coherent, non-corrupted tissue region.
For a given WSI graph $\mathcal{G}_i = (V_i, \mathcal{E}_i, X_i)$, we generate a ``corrupted" feature set $\mathbf{\tilde{X}}$ via random permutation. 
Let $\pi: V_i \to V_i$ be a random permutation function. 
The corrupted feature for node $j$ is:
\begin{equation}
\tilde{x}_{ij} = x_{i, \pi(j)}.
\end{equation}
Crucially, the structural topology $\mathcal{E}_i$ (the edges) remains fixed. 
This creates a mismatch where the physical location (defined by edges) receives the wrong texture (defined by $\tilde{x}$).

\paragraph{Shared Encoding.}
We employ a Siamese architecture where the \emph{same} GNN encoder $g_\nu$ processes both the real and corrupted graphs. 
This ensures the gradients update the main feature extractor:
\begin{align*}
h_{ij}^{\text{real}} &= g_\nu(x_{ij}, \mathcal{N}_{ij}) \quad &\text{(Coherent Context)} \\
h_{ij}^{\text{fake}} &= g_\nu(\tilde{x}_{ij}, \mathcal{N}_{ij}) \quad &\text{(Mismatched Context)}
\end{align*}
Note that $h{ij}^{\text{fake}}$ is computed using the \emph{original} neighbors of node $j$, but with the \emph{shuffled} center node feature (or vice versa, depending on implementation; shuffling the whole feature matrix $\mathbf{X}$ is computationally most efficient).\paragraph{Auxiliary Objective.}The auxiliary objective is to maximize the mutual information between the patch and its context. We minimize the Binary Cross-Entropy (BCE) loss to distinguish real embeddings from imposters:\begin{equation}\mathcal{L}^{\text{consistency}}{(i)} = - \frac{1}{N_i} \sum_{j=1}^{N_i} \Big[ \underbrace{\log D_\psi(h_{ij}^{\text{real}})}_{\text{Identify True Context}} + \underbrace{\log (1 - D\psi(h_{ij}^{\text{fake}}))}_{\text{Reject Imposter}} \Big].
\end{equation}
Here, $D\psi(\cdot)$ is a simple projection head (e.g., a linear layer followed by a sigmoid).

To minimize this loss, the encoder $g_\nu$ must learn to aggregate neighbor information. If $g_\nu$ were to ignore the neighbors (acting like a standard MLP), then $h_{ij}^{\text{fake}}$ would look statistically identical to a ``real" embedding from another part of the slide, making them indistinguishable. By forcing the discriminator to separate them, $g_\nu$ is compelled to encode the correlation between a center patch and its neighbors, effectively capturing local tissue morphology.



\section{Experiment Details}\label{app:experiment_details}

\paragraph{Implementation Details}
\begin{itemize}
    \item Feature Extractor: We utilized the UNI vision foundation model to extract 1024-dimensional features for each patch otherwise specified. 
    For Table~\ref{tab:auc_comparison_across_feature_types}, we vary the choice of vision foundational model with an intent to measure the feature effect. 
    \item Graph: $k$-NN graph with $k=50$, weighted by distance ($\sigma^2$ set to the maximum nearest neighbor distance).
    \item Network: 2-layer GAT with 256 hidden units, followed by ABMIL pooling.
    \item Training: Adam optimizer, learning rate 1e-4, weight decay 1e-5. The Jigsaw task used a $G = 10$, with random masking rate being 10\%.
    \item Evaluation: 3-fold cross-validation for the experiments shown in Table~\ref{tab:auc_comparison_across_cancer_types}, and 10-for cross-validation for Table~\ref{tab:auc_comparison_across_feature_types}. 
    We report Area Under the ROC Curve (AUC) of test set.
\end{itemize}



\paragraph{Data Preprocessing for TCGA-BRCA.}

The dataset contains a total of 1,133 whole-slide images (WSIs) corresponding to 1,062 patients. 
Among these, 1,009 slides from 948 patients are annotated with a binary histological label (Ductal or Lobular) while the remaining slides are labeled as mixed and excluded from further analysis.

The binary-labeled subset is class-imbalanced, comprising 805 Ductal slides and 204 Lobular slides. 
Because some patients contribute multiple slides, we restrict the analysis to a single slide per patient to avoid patient-level information leakage. 
After removing patients with more than one slide, the final dataset used in our experiments consists of 892 slides, including 710 Ductal and 182 Lobular cases.

\begin{table*}[ht]
    \centering
    \begin{tabular}{lcccc}
        \toprule
        \textbf{Method} & \textbf{UNI} & \textbf{gigapath} & \textbf{retccl} & \textbf{Virchow2}\\
        \midrule
        Embedding dim & 1024 & 1536 & 2048 & 2560 \\
        Num of patches (avg)& 3278.1 & 1310.5 & 1033.1 & 3287.1 \\
        Num of patches (min)& 50 & 11 & 11 & 50 \\
        Num of patches (max)& 12050 & 6631 & 5313 & 12050 \\
        \bottomrule
    \end{tabular}
    \caption{Summary statistics fo TCGA-BRCA dataset across four different foundational embeddings.}
    \label{tab:tcga_brca_data_summary}
\end{table*}

\section{Additional Dataset Descriptions}
\label{app:datasets}

All experiments use patch-level features extracted with UNI2 \citep{chen2024towards}, yielding 1536-dimensional embeddings per patch. We evaluate on the following datasets:

\begin{itemize}
\item \textbf{TCGA-NSCLC.} Binary subtype classification (LUAD vs.\ LUSC) using preprocessed UNI2 embeddings\footnote{https://huggingface.co/datasets/MahmoodLab/UNI2-h-features/tree/main/TCGA} from MahmoodLab~\citep{chen2024towards}.
The followings are all UNI 2 embedding with 1536 dimension.

\item \textbf{TCGA-COAD.} Binary MSI status prediction (MSI-H vs.\ MSS). Labels are derived from MSIsensor scores obtained via \href{https://www.cbioportal.org/study/clinicalData?id=coadread_tcga_pan_can_atlas_2018}{cBioPortal}, with a threshold of $\geq 10$ defining MSI-H~\citep{middha2017reliable}.

\item \textbf{TCGA-HNSC.} Binary HPV status prediction (HPV$+$ vs.\ HPV$-$) using clinical annotations from  
\href{https://www.cbioportal.org/study/summary?id=hnsc_tcga_pan_can_atlas_2018}{cBioPortal}

\item \textbf{PANDA.} Ordinal Gleason grading into ISUP grades 0--5 across 10{,}615 prostate biopsy WSIs~\citep{bulten2022artificial}.

\item \textbf{CAMELYON16.} Binary detection of lymph node metastasis in breast cancer (tumor vs.\ normal) across 399 WSIs (239 normal, 160 tumor), using preprocessed UNI2 features\footnote{https://huggingface.co/datasets/kaczmarj/camelyon16-uni}. 
\end{itemize}

\section{Additional Experiments}
\subsection{Ablation Studies on varying the graph size $k$}
Table~\ref{tab:auc_comparison_k_ablation} reports the effect of $k$-NN graph connectivity on a graph-only model with ABMIL aggregation, without jigsaw regularization.
The results reveal dataset-dependent sensitivity to $k$ with no single value dominating across cohorts.
BRCA and COAD exhibit opposing trends: BRCA peaks at $k=50$ (AUC $0.930$) and degrades with sparser graphs, whereas COAD favors $k=24$ (AUC $0.831$) and drops sharply at $k=8$($0.775$, standard deviation $\pm 0.145$), suggesting that overly local neighborhoods fail to capture the spatial context relevant for MSI prediction.
HNSC shows a consistent decline as $k$ increases, peaking at $k=8$ (AUC $0.855$), which may reflect the utility of strictly local tissue interactions for HPV status prediction.
Overall, $k=50$ offers the most stable performance across cohorts and is adopted as the default in all other experiments.

\begin{table}[!t]
\caption{Effect of $k$-NN graph connectivity on classification performance. 
All models use spatial-GAT convolution with ABMIL aggregation, without jigsaw regularization. Mean $\pm$ std AUC over 3-fold cross-validation is reported.}
\label{tab:auc_comparison_k_ablation}
\centering
\resizebox{0.6\linewidth}{!}{%
\begin{tabular}{@{}cccc@{}}
\toprule
\textbf{$k$} & \textbf{BRCA}   & \textbf{HNSC}   & \textbf{MSI}    \\ \midrule
8         & 0.914 $\pm$  0.029  & 0.855 $\pm$  0.056 & 0.775 $\pm$  0.145 \\
24        & 0.893 $\pm$  0.103 & 0.832 $\pm$  0.069 & 0.8305 $\pm$  0.0931 \\
50        & 0.930 $\pm$  0.055 & 0.821 $\pm$ 0.067  & 0.815 $\pm$  0.113 \\
\bottomrule
\end{tabular}
}
\end{table}

\subsection{Ablation Studies on the choice of graph convolutions}
To isolate the effect of graph convolution choice, we fix all other design 
decisions: we use a graph-enhanced MIL architecture \emph{without} jigsaw 
regularization, with graphs constructed as $k{=}50$ nearest-neighbor graphs 
using Gaussian edge weights.
We compare five aggregation strategies: ABMIL (no-graph baseline), (2) Spatial-GAT, (3) GCN, (4) GIN, and (5) GraphSAGE, across three clinically distinct tasks.
Performance is reported as mean $\pm$ std AUC over cross-validation folds in 
Table~\ref{tab:auc_comparison_graph_conv_ablation}.

\begin{enumerate}
    \item \textbf{Spatial-GAT is the most robust graph convolution.}
    GAT achieves the highest AUC on BRCA ($0.9304$) and ranks second on HNSC 
    ($0.8206$) and COAD ($0.8152$), making it the most consistent performer among graph-based methods across datasets.
    \item \textbf{Graph structure helps substantially on some tasks but not 
    universally.}
    On BRCA, all four GNN variants except GCN outperform or match ABMIL, with 
    GAT improving over ABMIL by over four points in mean AUC while halving the 
    standard deviation ($0.0554$ vs.\ $0.1138$), indicating both stronger and 
    more stable performance.
    On HNSC, GCN ($0.8325$) and GAT ($0.8206$) both outperform ABMIL 
    ($0.8094$).
    On COAD, however, only GIN ($0.8327$) exceeds the ABMIL baseline 
    ($0.8223$), while GAT ($0.8152$) and the remaining GNN variants fall 
    below it.
    This suggests that the benefit of explicit spatial modeling depends on 
    whether the discriminative signal is spatially distributed or predominantly 
    local.
    \item \textbf{The optimal convolution is dataset-dependent.}
    GCN leads on HNSC and GIN on COAD, reflecting that different cancer types 
    exhibit distinct spatial organization in their cellur spatial organization.
    MSI status, which is strongly associated with local morphological features 
    such as tumor-infiltrating lymphocytes and mucinous differentiation, may 
    favor GIN's injective neighborhood aggregation, which captures fine-grained 
    local structural motifs.
    \item \textbf{Spatial-GAT is biologically motivated.}
    Unlike GCN, which aggregates neighbors with fixed normalization, or GIN, 
    which captures structural equivalence, GAT's learned attention weights 
    reflect the functional heterogeneity of tissue; adaptively upweighting 
    spatially proximal patches that are contextually relevant.
    This aligns with how pathologists interpret tissue architecture and makes 
    the model's spatial reasoning directly interpretable through attention maps.
    We therefore adopt Spatial-GAT as the default backbone for subsequent 
    experiments.
\end{enumerate}

\begin{table}
\caption{Ablation over graph convolution type on small TCGA benchmarks. 
All methods use $k{=}50$ nearest-neighbor graphs with Gaussian edge weights 
and no jigsaw regularization. 
\textbf{Bold}: best per column; \underline{underline}: second best. 
3-fold cross-validated test ROC-AUC is reported (mean $\pm$ std).}
\label{tab:auc_comparison_graph_conv_ablation}
\centering
\resizebox{0.8\linewidth}{!}{%
\begin{tabular}{@{}cccc@{}}
\toprule
     & \textbf{BRCA}                     & \textbf{HNSC}                    & \textbf{COAD}                      \\ \midrule
Spatial-GAT (Ours)  & \textbf{0.9304 $\pm$  0.0554} & \underline{0.8206 $\pm$  0.067}    & \underline{0.8152 $\pm$  0.1131}    \\
GCN \citep{kipf2016semi} & 0.8247 $\pm$  0.0909          & \textbf{0.8325 $\pm$  0.095} & 0.8091 $\pm$  0.1298          \\
GIN \citep{xu2018powerful} & 0.8394 $\pm$  0.0425          & 0.796 $\pm$  0.0654          & \textbf{0.8327 $\pm$  0.1074} \\
SAGE \citep{hamilton2017inductive}& \underline{0.8774 $\pm$  0.056}     & 0.8163 $\pm$  0.0842         & 0.7775 $\pm$  0.105           \\\midrule
ABMIL \citep{ilse2018attention}& 0.8844 $\pm$ 0.1138 & 0.8094 $\pm$ 0.0615 & 0.8223 $\pm$ 0.0853 \\ \bottomrule
\end{tabular}}
\end{table}

\end{document}